\pdfoutput=1

\documentclass[11pt]{article}

\usepackage{ACL2023}

\usepackage{times}
\usepackage{latexsym}
\usepackage{pifont}
\usepackage{newunicodechar}
\usepackage{algorithm}
\usepackage{multirow}
\usepackage{array}
\usepackage{subcaption}
\usepackage{bbm}
\usepackage{enumitem}
\usepackage{booktabs}
\usepackage[noend]{algpseudocode}
\newunicodechar{✓}{\ding{51}}
\newunicodechar{✗}{\ding{55}}

\usepackage[T1]{fontenc}

\usepackage[utf8]{inputenc}

\usepackage{microtype}
\usepackage{amsmath}
\usepackage{amssymb}
\usepackage{graphicx}

\usepackage{inconsolata}

\title{Explore Spurious Correlations at the Concept Level in Language Models for Text Classification} 

\author{Yuhang Zhou$^{\ 1}$, Paiheng Xu$^{\ 2}$, Xiaoyu Liu$^{\ 2}$, Bang An$^{\ 2}$, Wei Ai$^{\ 1}$, Furong Huang$^{\ 2}$\\
         $^{1}$ College of Information Studies, University of Maryland, College Park \\ $^{2}$ Department of Computer Science, University of Maryland, College Park \\ \texttt{\{tonyzhou, paiheng, xliu1231, bangan, aiwei, furongh\}@umd.edu}}

\begin{document}
\maketitle
\begin{abstract}
Language models (LMs) have achieved notable success in numerous NLP tasks, employing both fine-tuning and in-context learning (ICL) methods. While language models demonstrate exceptional performance, they face robustness challenges due to spurious correlations arising from imbalanced label distributions in training data or ICL exemplars. Previous research has primarily concentrated on word, phrase, and syntax features, neglecting the concept level, often due to the absence of concept labels and difficulty in identifying conceptual content in input texts. This paper introduces two main contributions. First, we employ ChatGPT to assign concept labels to texts, assessing concept bias in models during fine-tuning or ICL on test data. We find that LMs, when encountering spurious correlations between a concept and a label in training or prompts, resort to shortcuts for predictions. Second, we introduce a data rebalancing technique that incorporates ChatGPT-generated counterfactual data, thereby balancing label distribution and mitigating spurious correlations. Our method's efficacy, surpassing traditional token removal approaches, is validated through extensive testing. \footnote{Our code and data are available at \url{https://github.com/Tonyzhou98/concept-spurious-correlation}}


\end{abstract}

\section{Introduction}
Pre-trained language models (LMs), leveraging extensive text corpora in their pre-training phase, have demonstrated remarkable effectiveness in a variety of natural language understanding tasks \cite{wei2022chain, devlin2018bert}. Nevertheless, LMs encounter issues with spurious correlations during fine-tuning or instruction-following stages \cite{zhang2022correct, wang-etal-2022-identifying, tang2023large}. These correlations involve specific associations between features and labels that, while prevalent in training data, are erroneously generalized as rules, leading to reduced performance.

\begin{figure}
     \centering
     \includegraphics[width=\linewidth]{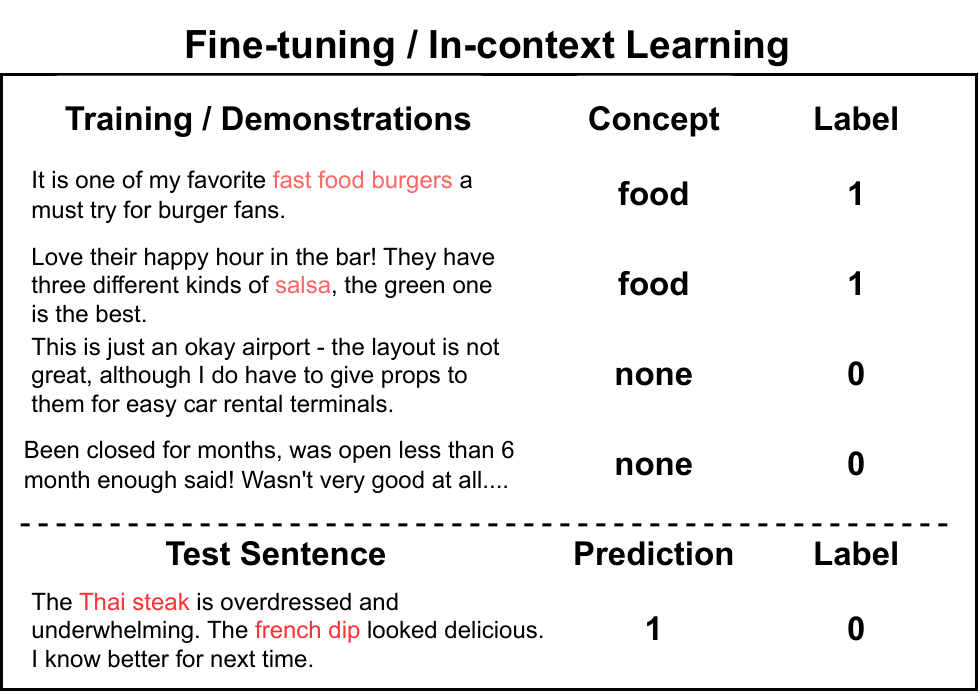}
     \caption{\label{fig:concept_example} \textbf{Example of concept-level spurious correlations.} In the training data or demonstrations, texts containing the concept ``food'' are mostly with label 1 (positive sentiment). During test, when encountering a sentence with the tokens ``Thai steak,'' not appearing in the training/prompts but indicating the concept ``food'', the models rely on the shortcut between the concept ``food'' and label 1 to give the wrong prediction.}
\end{figure}


Current research on spurious correlations in LMs spans various dimensions, such as token-level shortcuts in text classification \cite{wang-etal-2022-identifying, tang2023large, chew2023understanding}, syntactic heuristics in natural language inference \cite{mccoy2019right}, sentence triggers in text classification \cite{tang2023large, jia2017adversarial}, and topic shortcuts in machine translation \cite{borah2023measuring}. Moreover, spurious correlations with demographic concepts like race or sex, raise fairness concerns \cite{kleinberg2018algorithmic}. Yet, studies seldom address semantic spurious correlations at a broader concept level.


We define spurious correlations at the concept level as: Most texts featuring a certain concept in training data (or prompts) are linked with a specific label, leading LMs to inappropriately rely on this association for predictions. For instance, in Figure \ref{fig:concept_example}, terms like ``salsa,'' ``fast food burgers,'' or ``Thai steak'' denote the concept ``food.'' A prevalent association between ``food'' and label 1 in training data or prompts results in LMs forming a concept-level spurious correlation, mistakenly assigning some ``food''-related texts to label 1.

\begin{figure}
     \centering
     \includegraphics[width=\linewidth]{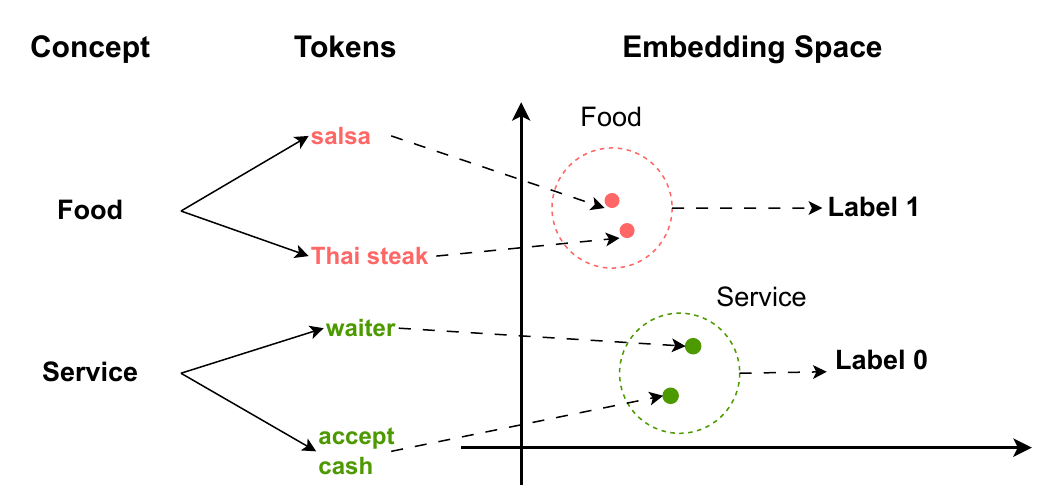}
     \caption{\label{fig:concept_reason} A concept can be expressed in multiple expressions, and in the embedding space of LMs, these expressions of one concept can be mapped into similar positions. LMs will form a shortcut between a specific concept and a label and utilize in the future prediction.}
\end{figure}

The tendency of LMs to learn concept-level shortcuts might stem from the formation of similar embeddings for expressions related to the same concept during fine-tuning or pre-training, driven by their semantic similarities. 
As Figure \ref{fig:concept_reason} suggests, various expressions of a concept cluster closely in the embedding space of fine-tuned or pre-trained LMs. 
When similar embeddings frequently coincide with a label in training or demonstrations, LMs tend to adopt the shortcut. 
We offer an in-depth analysis using a specific dataset in Section \ref{sec:clustering}.


In the first part of our study, we assess and quantify concept-level spurious correlations in LMs across both fine-tuning and ICL scenarios within text classification tasks. 
Initially, we employ the advanced large language model (LLM), ChatGPT, to identify relevant concepts in each dataset \cite{ouyang2022training} and to predict the presence of these concept labels. 
In the fine-tuning setting, we train LMs on both the original dataset and a concept-biased counterpart. 
Our findings indicate that LMs exhibit concept-level spurious correlations in standard benchmarks, with more pronounced prediction biases emerging from increasingly imbalanced data. 
In the ICL setting, we compare the performance of LMs on concept-balanced and concept-biased prompts, demonstrating that biased prompts lead to more skewed inferences.

The second part of the paper explores the use of data rebalancing techniques to counteract these spurious correlations in a fine-tuning framework. 
We introduce an upsampling strategy that incorporates counterfactual texts generated by ChatGPT, which effectively reduces bias while maintaining the utility (i.e., accuracy) of the LMs.
In summary, our research makes three significant contributions:
\vspace{-\topsep}
\begin{itemize}[leftmargin=*]
    \setlength\itemsep{-0.5em}
    \item We are the first to 
    investigate spurious correlations at a general concept level and introduce a metric to quantify these correlations.
    
    \item 
    Through experiments on various benchmark data for text classification, we demonstrate that LMs are prone to adopting learned concept-level shortcuts in both fine-tuning and ICL settings.

    \item 
    We introduce an effective upsampling approach, incorporating counterfactuals generated by LLMs, to mitigate concept-level bias.

\end{itemize}
\section{Exploring Concept-level Spurious Correlations}

\subsection{Obtaining the Concept Labels}
\label{sec:obtain_concept}
Due to the lack of human-annotated metadata indicating concepts in most text classification datasets, and considering the superior capabilities of LLMs in text annotation tasks over human annotators \cite{gilardi2023chatgpt}, we utilize ChatGPT (GPT-3.5) to annotate concept labels for sentences in text classification datasets \cite{ouyang2022training}. 
Our annotation process involves an annotation prompt $P_a$ that contains the annotation instruction and five demonstrations, a text input $x$, LLM $M_a$, and a candidate concept set $C = \{C_1, C_2, \cdots, C_k\}$ (we describe how we curate the candidate set in Section \ref{sec:experiments}).

The annotation process is formalized as: $a(x) = M_a(P_a \| C \| x)$, where $a(x)$, the set of concept labels for text $x$, may contain zero or several concepts selected from the pre-defined concept set $C$ ($a(x) \subset C$), and $\|$ denotes the concatenation operation. 
To ensure reliability, we repeat the annotation process twice with a temperature setting of 0.7 and retain only those examples and labels that are consistently identified by both LLM annotators.

\subsection{Measuring Concept Spurious Correlations}
For the text classification task, we consider an input $x \in \mathcal{X}$ 
accompanied by concept labels $a(x) \subset C$.
Each input is associated with a ground truth classification label $y = l$ from the output label space $\mathcal{Y}$, $l \in \{0, 1, \cdots, n\}$.
Given a LM classifier $M: \mathcal{X} \rightarrow \mathcal{Y}$, if the model avoids utilizing potential concept-level shortcuts from $c \rightarrow y, c\in C$, the following condition is satisfied:
\vspace{-1em}
\begin{align}
\label{eq:1}
    &\mathbb{E}_x[p_M(\hat{y} = l | x, c\in a(x), y = l)] \\
    = &\mathbb{E}_{x'}[p_M(\hat{y} = l | x', c \notin a(x'), y = l)] \quad \forall l\in\mathcal{Y} \notag.
\end{align}
Here, $\hat{y}$ denotes the predicted label, while $p_M$ represents the probability predicted by model $M$. The inputs $x$ and $x'$, belonging to the space $\mathcal{X}$, contain the concept $c$ or do not contain it, respectively.

Equation \ref{eq:1} implies a critical condition: regardless of the presence of concept $c$ in the input, the models should maintain an unbiased estimate of the predicted probability on average. The expression $\mathbb{E}_{x}[p_M(\hat{y} = l | x, c \in a(x), y = l)]$ can be interpreted as the model's accuracy on texts that are labeled $l$ and incorporate the concept $c$.



Denote $\Delta_{c_i}$ as the difference in model accuracy between texts with or without concept $c$ that have label $i \in \mathcal{Y}$. We further infer from Equation \ref{eq:1} that:
\begin{align}
    \Delta_{c_i} = & \mathbb{E}_x[p_M(\hat{y} = i | x, c, y = i)] \\
    - &\mathbb{E}_{x'}[p_M(\hat{y} = i | x', \neg c, y = i)] = 0 \notag,
\end{align}
where $\neg c$ denotes concept $c$ is not in input $x$. 
We hypothesize, if there exists a spurious correlation in models between concept $c$ and label $i$, the following conditions would hold:
\begin{equation*}
\resizebox{\columnwidth}{!}{%
$
\begin{aligned}
    & \mathbb{E}_x[p_M(\hat{y} = i | x, c, y = i)] > \mathbb{E}_{x'}[p_M(\hat{y} = i | x', \neg c, y = i)] \\
    & \mathbb{E}_x[p_M(\hat{y} = j | x, c, y = j)] < \mathbb{E}_{x'}[p_M(\hat{y} = j | x', \neg c, y = j)]
\end{aligned}
$
}
\end{equation*}
\vspace{-0.3em}
Then we have $\Delta_{c_i} > 0 > \Delta_{c_j}$. Otherwise, if the spurious correlation is between $c$ and $j$, then $\Delta_{c_j} > 0 > \Delta_{c_i}$. We propose to measure the discrepancy between $\Delta_{c_i}$ and $\Delta_{c_j}$ to quantify the spurious correlation.
Hence, considering the output space $\mathcal{Y}$, we quantify the model's reliance on shortcut mapping as the average discrepancy in the accuracy difference $\Delta_{c_i} - \Delta_{c_j}$ across all label combinations. 
\begin{equation*}
    \text{Bias@C} = \frac{1}{\binom{n}{2}} \sum_{i, j \in \mathcal{Y}} (\Delta_{c_i} - \Delta_{c_j}), i > j
\end{equation*}
For the binary classification task, the bias measurement is simplified to $\text{Bias@C} = \Delta_{c_1} - \Delta_{c_0}$

A Bias@C approaching 0 indicates minimal reliance on concept shortcuts.
Conversely, a positive Bias@C value suggests that model is more likely to predict larger labels when the input includes concept $c$,  and the opposite for a negative value.


\subsection{Evaluation of Model Robustness to Concept Shortcut in Fine-tuning}
\label{sec:dataset_construct}

To assess LMs' robustness against spurious correlations for concept $c$ across varying scales of concept bias during fine-tuning, we fine-tune models on the original dataset $\mathcal{D}_{ori}$ and a concept-biased dataset $\mathcal{D}^c_{biased}$ separately. 
To further demonstrate the impact of concept-level spurious correlation, we construct $\mathcal{D}^c_{biased}$ of concept $c$ by filtering $\mathcal{D}_{ori}$,
where, for each data point, we only keep those with the majority labels under concept $c$.
After fine-tuning on $\mathcal{D}_{ori}$ or $\mathcal{D}^c_{biased}$, we evaluate models on test data using Bias@C to quantify spurious correlations.

We report accuracy on the test data for utility performance. However, label distributions with or without the concept $c$ may be imbalanced. 
Following previous work \cite{chew2023understanding}, we rebalance the test set by downsampling and report the inference accuracy (\textbf{robust accuracy}) on the balanced subset for examples with concept $c$ (\textbf{Acc@C}) and without concept $c$ (\textbf{Acc@NoC}), respectively.

\subsection{Evaluation of Model Robustness to Concept Shortcut in ICL}
\label{sec:prompt_construction}
As LLMs have shown outstanding performances with the ICL setting, we are interested in investigating the concept shortcut in the demonstrations. The prompt $P$ for ICL contains three parts: 1) the instruction $s$, 2) the demonstrations with $h$ exemplars (text + labels), and 3) the test input $x_{test}$. 

We consider the sentiment classification task and concatenate the $h$ exemplars together with the form ``Input: $x$. The sentiment label is $v(y)$''. 
The label verbalizer $v(y)$ will transfer $0$ to ``negative'' and $1$ to ``positive'' when the label is binary and will maintain the original numerical rating scales when multiple classes ($n\geq3$). The ICL process is formulated as $f(x_{test}) = M(P \| x_{test})$, where $f(x_{test})$ is a categorical variable belonging to $\mathcal{Y}$. 

We create two types of prompts: the biased prompt $P_{biased}$ and the balanced prompt $P_{balanced}$ by changing the label distributions in the demonstrations. For $P_{biased}$, we insert $\frac{h}{2}$ numbers of exemplars containing the concept $c$ with the label $l \in \{l_1, l_2, \cdots, l_k\}$ (the majority ground truth labels) and $\frac{h}{2}$ numbers of exemplars without $c$ with the other labels. For $P_{balanced}$, we split the exemplars with concept $c$ or not half by half, but ensure balanced labels. We compare the results of Bias@C and robust accuracy with two types of prompts. 
Since the ICL are sensitive to the exemplars, we repeat the experiments three times with differently selected exemplars and report the average values.

\begin{table}[t]
\small
\centering
\resizebox{\columnwidth}{!}{%
\begin{tabular}{lcccl}
\hline
Dataset     & \# Training   & \# Test   & \# Labels & Concept                 \\ \hline
AS          &  70,117             & 8,000     & 5         & size, color, style      \\ \hline
IMDB        &  14,956             & 4,000     & 2         & acting, comedy, music   \\ \hline
Yelp        &  34,184             & 4,000     & 2         & food, price, service    \\ \hline
CeBaB       &  7,350         & 2,000     & 5         & service, food, ambiance \\ \hline
BoolQ       &  2393         & 2,000     & 2         & country, history \\ \hline
\end{tabular}
}
\caption{\label{table:dataset} Dataset statistics and the labeled concept for each dataset. AS represents the Amazon Shoe dataset.}
\end{table}

\section{Dataset Construction and Analysis}
\label{sec:experiments}

\noindent \textbf{Models} \quad
We assess and mitigate concept-level bias in DistilBERT and LLAMA2 7B in fine-tuning setting \cite{sanh2019distilbert, touvron2023llama} and  GPT3.5 in the ICL setting \cite{ouyang2022training}. We fully fine-tune the DistillBert. For LLAMA2, we apply the Lora method for efficient fine-tuning \cite{hu2021lora}. Details of the model implementations are included in Appendix \ref{sec:implementation}.

\vspace{0.1cm}
\noindent \textbf{Dataset} \quad
We select four sentiment classification tasks to evaluate the model robustness: Yelp \cite{zhang2015character}, IMDB \cite{maas2011learning}, Amazon Shoe \cite{he2016ups}, and CeBaB \cite{abraham2022cebab}. Amazon shoe and CeBaB datasets with 5 classes, 0 indicating the most negative and 4 indicating the most positive, are reviews of shoes in Amazon and OpenTable. IMDB and Yelp are binary classification datasets (0 indicating negative and 1 indicating positive), with reviews from the IMDB and Yelp platforms. 

Additionally, we include a binary question answering (QA) dataset BoolQ \cite{clark2019boolq}, which asks Yes/No questions. It takes a paired question and passage as the input to LMs and outputs 1 (Yes) or 0 (No). In the following part, we define the \textbf{positive} class as datapoints with Label 3 and 4 for the 5-way classification tasks and those with Label 1 for the binary classification task. We define the remaining datapoints as the \textbf{negative} class. 

\vspace{0.1cm}
\noindent \textbf{Concept} \quad
For CeBaB, we adopt human-annotated concept labels. For Amazon Shoe, IMDB, Yelp, and BoolQ where there are no concept annotations, we first use ChatGPT to query the concepts embedded in each sentence and count the number of occurrences for each concept following \cite{fang2022data} to generate concept-level explanations. We then extract the most frequent concepts and identify the concepts whose existence should not influence the sentiments of the text or the Yes/No answer to the question (2 concepts for BoolQ due to more diverse topics and 3 concepts for other datasets). Finally, we use ChatGPT to annotate whether each text input contains the selected concept. 

To examine the quality of the annotation, we experiment on the human-annotated ``service'' concept from the CeBaB dataset and ask the ChatGPT to label the concept again. We find that ChatGPT can achieve an accuracy of 90.4\% to the gold standard concept labels, comparable to an average agreement of 92.9\% for five human annotators given by CeBaB, indicating the reliability of LLM annotations. 
Table \ref{table:dataset}, 
presents dataset statistics and the labeled concept lists for the five datasets.

\subsection{Biased Dataset Construction}

\begin{figure}[!t]
     \centering
     \includegraphics[width=0.9\linewidth]{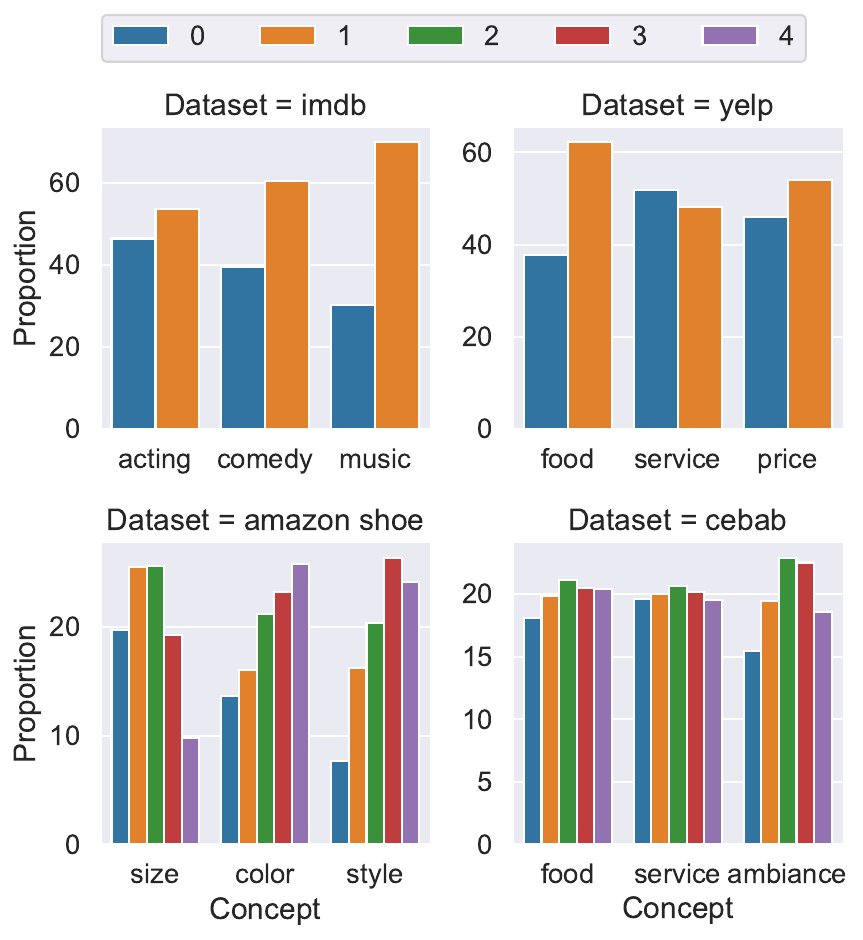}
     \caption{\label{fig:label_dist} \textbf{Label distribution of the texts with a specific concept for each dataset.} We can observe the label distribution in multiple concepts, such as ``music'' in IMDB, ``food'' in Yelp datasets are highly imbalanced.}
\end{figure}

We first visualize the label distribution for the input texts with the concept $c$ for each sentiment classification in Figure \ref{fig:label_dist}.
We observe that for the original datasets, the concept-label distributions are balanced for most concepts, but not as balanced for concepts such as ``food'' in Yelp dataset, ``music'' in IMDB, and ``style'' in Amazon Shoe. In 10/12 cases, positive labels comprise large proportions of the corpus with certain concepts. 
To further demonstrate the impact of concept-level spurious correlation caused by imbalanced concept-label distribution, we construct a biased dataset $\mathcal{D}^c_{biased}$ which, for each concept $c$, only includes the majority class (positive or negative). Specifically, we keep negative class for ``size'' in Amazon Shoe and ``service'' in Yelp. 
For other concepts in the sentiment datasets, we keep positive class. 
For BoolQ, we keep negative class with ``country'' and positive class with ``history''.

\subsection{Embedding Analysis of Associated Tokens}
\label{sec:clustering}

As shown in Figure \ref{fig:concept_reason}, we hypothesize that expressions of a concept have similar semantic embeddings, leading to shortcut learning. To verify the hypothesis and further motivate the measurement of spurious correlations, we extract the embeddings of the associated tokens with each concept in the Amazon shoe dataset. We observe whether the embeddings of tokens associated with the same concept are similar using clustering.

We apply the point-wise mutual information (PMI) between the token and the concept to measure the association.
For a dataset with a concept $c$, we calculate the PMI of each token $t$ to concept $c$, which is $\text{PMI}(t, c) = \log \frac{p(t, c)}{p(t)p(c)}$, where $p(t)$, $p(c)$ and $p(t, c)$ refer to the probability of the text containing $t$, $c$ and both together. The higher value of PMI suggests a stronger association between $t$ and $c$. We present tokens with the top 10 PMI values for each concept in Table \ref{tab:top_pmi}.

\begin{table}[t]
\centering
\resizebox{0.95\columnwidth}{!}{%
\begin{tabular}{c|l}
\hline
Concept & Top associated tokens extracted from each concept                                                                           \\ \hline
Size    & 9m, small, c/d, sizing, 105, us, 95, 8w, 81/2, 7-75                                                                     \\ \hline
Color   & \begin{tabular}[c]{@{}l@{}}royal, camel, muted, champagne, color, taupe, \\ maroon, teal, greenish, white\end{tabular}  \\ \hline
Style   & \begin{tabular}[c]{@{}l@{}}stylish, vibe, comfort, swedish, look, all-time, \\ (55), styling, yearround, frumpy\end{tabular} \\ \hline
\end{tabular}
}
\caption{Tokens with high associations (top 10 PMI values) to each concept in Amazon Shoe dataset.}
\label{tab:top_pmi}
\end{table}

From the associated tokens in Table \ref{tab:top_pmi}, we observe tokens with various semantics associated with one concept, such as ``small,'' ``sizing'' and ``9m'' to express the ``size'' concept. 
We use the tokens in Table \ref{tab:top_pmi} to perform the clustering. We exclude tokens with special character, such as ``c/d,'' ``81/2'' and ``(55)'' that affect the interpretation of the results. We feed the tokens into the DistilBERT fine-tuned on the Amazon Shoe and retrieve the corresponding embedding from the model last layer. If the token is tokenized into multiple sub-words, we follow the previous work and calculate the average as the token embedding \cite{wolfe-caliskan-2021-low}. We calculate the cosine distance between their token embeddings and apply hierarchical clustering \cite{bar2001fast}. 

\begin{figure}[t]
\centering
    \includegraphics[width=0.9\linewidth]{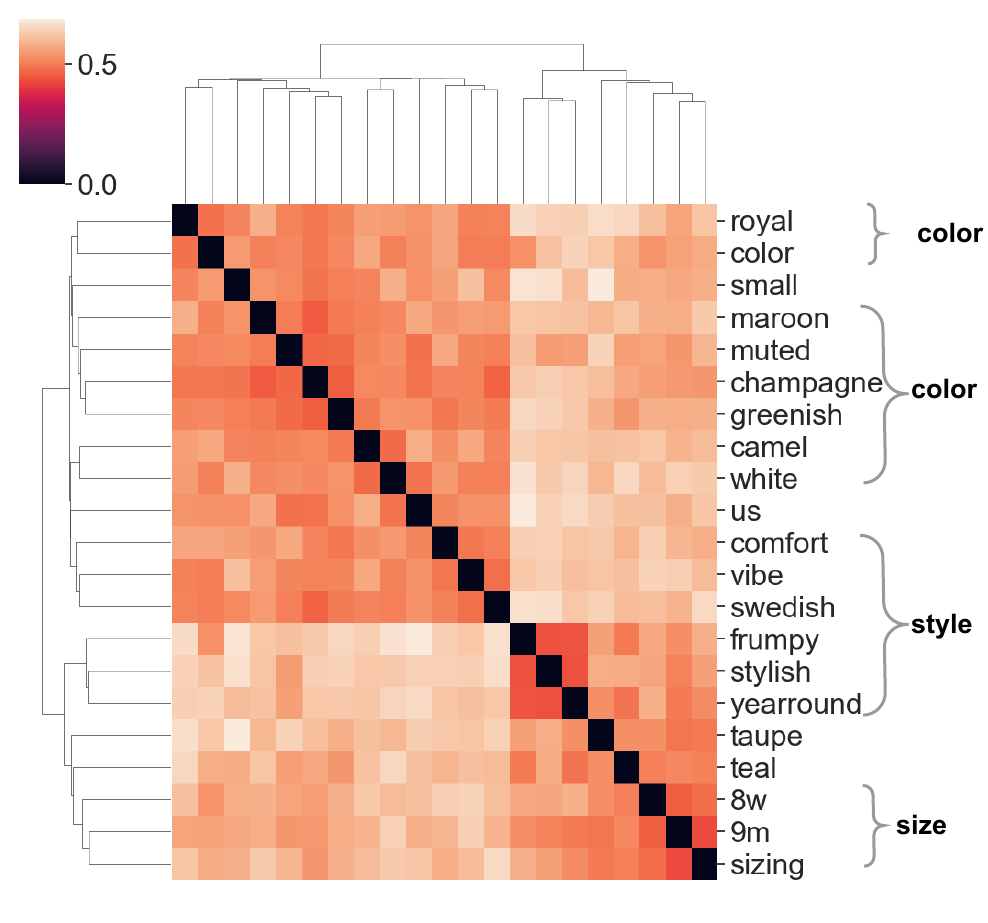}
    \caption{\textbf{Clusters of word embeddings of top associated tokens for each concept from Amazon shoe dataset.} 
    The dendrogram on the side indicates the hierarchical clustering structure among the tokens.}
    \label{fig:cluster_as}
\end{figure}

From Figure \ref{fig:cluster_as}, we can identify four small clusters, each representing a concept. We observe that the LMs will produce similar internal representations for tokens associated with the same concept label. If the label under a concept is imbalanced, the models may learn the undesired shortcut between similar embeddings and a label. This observation motivates the measurement of spurious correlation at the concept level.

\section{Results of Spurious Correlation Measurement}
\label{sec:main_result}

\begin{table*}[!htb]
\centering
\resizebox{\textwidth}{!}{%
\begin{tabular}{lccccccccc}
\hline
\multicolumn{1}{l|}{\multirow{2}{*}{Data: Amazon Shoe}} & \multicolumn{3}{c|}{Size (pos < neg)}                                             & \multicolumn{3}{c|}{Color (pos > neg)}                                            & \multicolumn{3}{c}{Style (pos > neg)}                        \\
\multicolumn{1}{l|}{}                        & Bias@C         & Acc@NoC        & \multicolumn{1}{c|}{Acc@C}          & Bias@C         & Acc@NoC        & \multicolumn{1}{c|}{Acc@C}          & Bias@C         & Acc@NoC        & Acc@C          \\ \hline
\multicolumn{1}{l|}{Trained on $\mathcal{D}_{ori}$}       & \textbf{2.11}  & \textbf{57.94} & \multicolumn{1}{c|}{\textbf{55.94}} & \textbf{1.38}  & 57.19          & \multicolumn{1}{c|}{\textbf{55.50}} & \textbf{11.56} & \textbf{57.18} & \textbf{56.12} \\
\multicolumn{1}{l|}{Trained on $\mathcal{D}^c_{biased}$}    & -3.77          & 56.75          & \multicolumn{1}{c|}{47.76}          & 14.99          & \textbf{57.56} & \multicolumn{1}{c|}{48.19}          & 13.74          & 56.39          & 54.92          \\ \hline
\multicolumn{1}{l|}{\multirow{2}{*}{Data: IMDB}} & \multicolumn{3}{c|}{Acting (pos > neg)}                                           & \multicolumn{3}{c|}{Comedy (pos > neg)}                                           & \multicolumn{3}{c}{Music (pos > neg)}                        \\
\multicolumn{1}{l|}{}                        & Bias@C         & Acc@NoC        & \multicolumn{1}{c|}{Acc@C}          & Bias@C         & Acc@NoC        & \multicolumn{1}{c|}{Acc@C}          & Bias@C         & Acc@NoC        & Acc@C          \\ \hline
\multicolumn{1}{l|}{Trained on $\mathcal{D}_{ori}$}       & \textbf{3.70}  & 88.49          & \multicolumn{1}{c|}{\textbf{91.24}} & \textbf{2.51}  & \textbf{91.62} & \multicolumn{1}{c|}{\textbf{91.85}} & 12.07 & \textbf{91.55} & 88.93          \\
\multicolumn{1}{l|}{Trained on $\mathcal{D}^c_{biased}$}    & 8.69           & \textbf{88.87} & \multicolumn{1}{c|}{88.67}          & 5.14           & 90.55          & \multicolumn{1}{c|}{90.50}          & \textbf{8.25}           & 90.55          & \textbf{89.30} \\ \hline
\multicolumn{1}{l|}{\multirow{2}{*}{Data: Yelp}} & \multicolumn{3}{c|}{Food (pos > neg)}                                             & \multicolumn{3}{c|}{Service (pos < neg)}                                          & \multicolumn{3}{c}{Price (pos > neg)}                        \\
\multicolumn{1}{l|}{}                        & Bias@C         & Acc@NoC        & \multicolumn{1}{c|}{Acc@C}          & Bias@C         & Acc@NoC        & \multicolumn{1}{c|}{Acc@C}          & Bias@C         & Acc@NoC        & Acc@C          \\ \hline
\multicolumn{1}{l|}{Trained on $\mathcal{D}_{ori}$}       & \textbf{3.42}  & 93.86          & \multicolumn{1}{c|}{97.09}          & \textbf{2.44}  & \textbf{93.33} & \multicolumn{1}{c|}{\textbf{97.23}} & \textbf{-0.32} & 94.04          & \textbf{94.44} \\
\multicolumn{1}{l|}{Trained on $\mathcal{D}^c_{biased}$}    & 3.93           & \textbf{93.98} & \multicolumn{1}{c|}{\textbf{97.58}} & -3.85          & 90.23          & \multicolumn{1}{c|}{95.14}          & 5.62           & \textbf{96.03} & 94.00          \\ \hline
\multicolumn{1}{l|}{\multirow{2}{*}{Data: CeBaB}} & \multicolumn{3}{c|}{Food (pos > neg)}                                             & \multicolumn{3}{c|}{Service (pos > neg)}                                          & \multicolumn{3}{c}{Ambiance (pos > neg)}                     \\
\multicolumn{1}{l|}{}                        & Bias@C         & Acc@NoC        & \multicolumn{1}{c|}{Acc@C}          & Bias@C         & Acc@NoC        & \multicolumn{1}{c|}{Acc@C}          & Bias@C         & Acc@NoC        & Acc@C          \\ \hline
\multicolumn{1}{l|}{Trained on $\mathcal{D}_{ori}$}       & \textbf{-0.71} & \textbf{69.48} & \multicolumn{1}{c|}{\textbf{74.12}} & \textbf{-0.34} & \textbf{69.70} & \multicolumn{1}{c|}{\textbf{74.45}} & \textbf{-0.90} & \textbf{72.62} & \textbf{75.68} \\
\multicolumn{1}{l|}{Trained on $\mathcal{D}^c_{biased}$}    & 14.94          & 61.17          & \multicolumn{1}{c|}{58.14}          & 13.06          & 67.08          & \multicolumn{1}{c|}{61.58}          & 7.02           & 69.60          & 65.96          \\ \hline
\end{tabular}
}
\caption{\textbf{Model fine-tuning performance with training on original dataset and concept biased dataset for four datasets.} Models trained on the original dataset $\mathcal{D}_{ori}$ tend to behave a bias in some concepts, where the label distribution under concepts is pretty uneven. When fine-tuned on the concept-biased dataset $\mathcal{D}^c_{biased}$, both bias (Bias@C) and accuracy results (Acc@C and Acc@NoC) suffer from performance drop. pos > neg: for this concept, more positive texts are in $\mathcal{D}_{ori}$ and in $\mathcal{D}_{biased}$, all texts containing this concept are positive, and vice versa for ``pos < neg''. The lower absolute values of Bias@C (smaller bias) and the higher accuracy values are in bold. 
}
\label{tab:finetune_main_results}
\end{table*}

\subsection{Spurious Correlations in Fine-tuning}
\label{sec:fine-tuning_results}


To evaluate the robustness to the concept shortcut in the fine-tuning setting, we fine-tune the models on the original dataset $\mathcal{D}_{ori}$ and the biased dataset $\mathcal{D}^c_{biased}$, respectively, and measure the concept bias. 
For each concept, we report the metric Bias@C to quantify the strength of spurious correlations and the robust accuracy for texts with and without concept, i.e., Acc@C and Acc@NoC, as the utility performance. For Bias@C, closer to 0 indicates weaker spurious correlations for concept $C$, and for robust accuracy, a higher value suggests better performance. We present the results for DistilBERT on sentiment classification in Table \ref{tab:finetune_main_results} and BoolQ dataset in Table \ref{tab:finetune_boolq} Appendix \ref{sec:supplementary}.

\noindent\textbf{Fine-tuned LMs present a clear concept bias when trained on both original and biased data.} \quad Table \ref{tab:finetune_main_results} shows that when the models are trained on $\mathcal{D}_{ori}$, they utilize spurious correlations in the training data to make inferences. For example, for ``style'' in the Amazon Shoe and ``music'' in IMDB, the Bias@C values in $\mathcal{D}_{ori}$ are large due to highly imbalanced label distribution.
Since these datasets are well curated and widely adopted, the fact that we are able to identify several highly biased concepts by only investigating the top 3 frequent concepts demonstrates the significance of spurious correlation.

Comparing the results between $\mathcal{D}_{ori}$ and $\mathcal{D}^c_{biased}$, we find that the absolute values of Bias@C are significantly higher when trained on $\mathcal{D}^c_{biased}$ in almost every concept, and the change direction of Bias@C is the same as the trend in the label distribution. For example, the value of Bias@C becomes negative for ``service'' in Yelp dataset, since we only keep negative reviews with the ``service'' in $\mathcal{D}^c_{biased}$.  These observations indicate that a greater bias in the fine-tuning dataset causes the model to rely more on spurious correlations to make predictions.

Regarding utility performance (Acc@C and Acc@NoC), we observe that the models trained on $\mathcal{D}^c_{biased}$ have a dramatic performance drop on the texts with the concept in most cases, and the average Acc@C decreases from 79.38\% to 74.31\%. This pattern suggests that larger spurious correlations affect the utility performance of fine-tuned models. Meanwhile, the average Acc@NoC drops from 78.08\% to 76.56\%. Its performance drop is not as large as the one of Acc@C, indicating that texts without the concept suffer less from the concept bias in the datasets.

Moreover, we find that for some concepts, the fine-tuned LMs suffer from severe spurious correlation, but the effect of the bias is not fully reflected in the difference between Acc@C and Acc@NoC. For example, the ``music'' concept in the IMDB dataset has Bias@C = 12.07\%, but the difference between Acc@C and Acc@NoC is less than 3\%. This is because if the model is biased towards one label due to the spurious correlation, the accuracy improvement towards the biased label can often offset the performance drop of the other side. 

We also verify that the concept bias is not simply due to the shortcut on a few words by masking out the associated tokens, and details are shown in Section \ref{sec:fine-tuning_mitigate_results}.
We show fine-tuning results of LLAMA2 models on $\mathcal{D}_{ori}$ and $\mathcal{D}^c_{biased}$ in Table \ref{tab:finetune_boolq} and \ref{tab:finetune_main_results_llama} (Appendix \ref{sec:supplementary}). Similar patterns suggest the generalizability of our findings on models of different sizes. 

\begin{table*}[!htb]
\centering
\resizebox{\textwidth}{!}{%
\begin{tabular}{lccccccccc}
\hline
\multicolumn{1}{l|}{\multirow{2}{*}{Data: Amazon Shoe}} & \multicolumn{3}{c|}{Size (pos \textless neg)}                        & \multicolumn{3}{c|}{Color (pos \textgreater neg)}                     & \multicolumn{3}{c}{Style (pos \textgreater neg)}    \\
\multicolumn{1}{l|}{}                        & Bias@C        & Acc@NoC        & \multicolumn{1}{c|}{Acc@C}          & Bias@C         & Acc@NoC        & \multicolumn{1}{c|}{Acc@C}          & Bias@C          & Acc@NoC         & Acc@C           \\ \hline
\multicolumn{1}{l|}{ICL with $P_{balanced}$}    & \textbf{0.95} & \textbf{50.37} & \multicolumn{1}{c|}{45.63}          & \textbf{9.46}  & \textbf{49.63} & \multicolumn{1}{c|}{\textbf{49.54}} & \textbf{11.58}  & \textbf{52.98}  & 53.91           \\
\multicolumn{1}{l|}{ICL with $P_{biased}$}      & -2.64         & 50.18          & \multicolumn{1}{c|}{\textbf{47.20}} & 10.99          & 49.58          & \multicolumn{1}{c|}{46.59}          & 12.56           & 50.21           & \textbf{54.35}  \\ \hline
\multicolumn{1}{l|}{\multirow{2}{*}{Data: IMDB}} & \multicolumn{3}{c|}{Acting (pos \textgreater neg)}                   & \multicolumn{3}{c|}{Comedy (pos \textgreater neg)}                    & \multicolumn{3}{c}{Music (pos \textgreater neg)}    \\
\multicolumn{1}{l|}{}                        & Bias@C        & Acc@NoC        & \multicolumn{1}{c|}{Acc@C}          & Bias@C         & Acc@NoC        & \multicolumn{1}{c|}{Acc@C}          & Bias@C          & Acc@NoC         & Acc@C           \\ \hline
\multicolumn{1}{l|}{ICL with $P_{balanced}$}    & \textbf{3.58} & 94.94          & \multicolumn{1}{c|}{\textbf{96.40}} & \textbf{3.70}  & \textbf{96.18} & \multicolumn{1}{c|}{\textbf{95.82}} & 6.40            & \textbf{95.43}  & \textbf{94.33}  \\
\multicolumn{1}{l|}{ICL with $P_{biased}$}      & 3.99          & \textbf{95.05} & \multicolumn{1}{c|}{96.17}          & 5.64           & 95.68          & \multicolumn{1}{c|}{94.71}          & \textbf{5.45}   & 95.33           & 94.26           \\ \hline
\multicolumn{1}{l|}{\multirow{2}{*}{Data: Yelp}} & \multicolumn{3}{c|}{Food (pos \textgreater neg)}                     & \multicolumn{3}{c|}{Service (pos \textgreater neg)}                   & \multicolumn{3}{c}{Price (pos \textgreater neg)}    \\
\multicolumn{1}{l|}{}                        & Bias@C        & Acc@NoC        & \multicolumn{1}{c|}{Acc@C}          & Bias@C         & Acc@NoC        & \multicolumn{1}{c|}{Acc@C}          & Bias@C          & Acc@NoC         & Acc@C           \\ \hline
\multicolumn{1}{l|}{ICL with $P_{balanced}$}    & 2.97          & \textbf{97.66} & \multicolumn{1}{c|}{98.21}          & \textbf{0.39}  & \textbf{97.78} & \multicolumn{1}{c|}{\textbf{98.60}} & \textbf{-0.87}  & 97.74           & 97.79           \\
\multicolumn{1}{l|}{ICL with $P_{biased}$}      & \textbf{1.70} & 97.54          & \multicolumn{1}{c|}{\textbf{98.84}} & 0.92           & 97.50          & \multicolumn{1}{c|}{98.58}          & 1.17            & \textbf{97.98}  & \textbf{98.46}  \\ \hline
\multicolumn{1}{l|}{\multirow{2}{*}{Data: CeBaB}} & \multicolumn{3}{c|}{Food (pos \textgreater neg)}                     & \multicolumn{3}{c|}{Service (pos \textless neg)}                      & \multicolumn{3}{c}{Ambiance (pos \textgreater neg)} \\
\multicolumn{1}{l|}{}                        & Bias@C        & Acc@NoC        & \multicolumn{1}{c|}{Acc@C}          & Bias@C         & Acc@NoC        & \multicolumn{1}{c|}{Acc@C}          & Bias@C          & Acc@NoC         & Acc@C           \\ \hline
\multicolumn{1}{l|}{ICL with $P_{balanced}$}    & \textbf{0.66}          & \textbf{61.06} & \multicolumn{1}{c|}{\textbf{64.35}} & \textbf{0.65} & 58.48          & \multicolumn{1}{c|}{62.08}          & 2.58   & \textbf{64.13}  & \textbf{64.96}  \\
\multicolumn{1}{l|}{ICL with $P_{biased}$}      & 3.24 & 54.55          & \multicolumn{1}{c|}{59.90}          & -2.77           & \textbf{61.91} & \multicolumn{1}{c|}{\textbf{65.53}} & \textbf{2.21}            & 61.39           & 62.45           \\ \hline
\end{tabular}
}
\caption{\textbf{Model ICL performance with prompting on balanced prompts $P_{balanced}$ and biased prompts $P_{biased}$.} Larger absolute values for Bias@C indicate that the concept-biased prompts enlarge the extend of models to rely on the shortcut in the demonstrations. The meaning of ``pos > neg'' and the values in bold are the same as in Table \ref{tab:finetune_main_results}.}
\label{tab:icl_main_results}
\end{table*}

\subsection{Spurious Correlations in ICL}
As LMs exhibit clear evidence of utilizing the concept shortcuts in the fine-tuning data, we also want to ask whether LMs use the shortcuts in the exemplars of the prompts when performing ICL. As discussed in Section \ref{sec:prompt_construction}, for each concept $c$ in five datasets, we construct a prompt with eight exemplars. 
Following a similar setup for fine-tuning, we only include the majority class (positive or negative) for exemplars with concept $c$. Specifically, for ``size'' in Amazon Shoe and ``service'' in Yelp, four exemplars with concept $c$ have negative labels and the other four without concept $c$ have positive labels. The labels are flipped for the rest of the concepts.
For the balanced prompt $P_{balanced}$, the label is evenly distributed for the exemplars with or without the concept. With the bias in label distribution for both texts with and without the concept, we expect the LM to use two types of shortcuts: a) from texts with the concept $c$ to one sentiment and b) from texts without the concept to the other sentiment.
We present the utility and bias results of ICL for sentiment classification dataset in Table \ref{tab:icl_main_results} and BoolQ in Table \ref{tab:finetune_boolq} (Appendix \ref{sec:supplementary}).

\noindent \textbf{Biased prompts enlarge the concept bias in ICL inference} \quad From Table \ref{tab:icl_main_results}, we observe a similar pattern for Bias@C as in the fine-tuning part. When the prompt changes from $P_{balanced}$ to $P_{biased}$, for ``service'' in Yelp and ``size'' in Amazon Shoe, where the exemplars with concept $c$ are negative, the values of Bias@C flip from positive to negative, and for most other concepts, where conceptual exemplars are all positive, the value of Bias@C increases. Furthermore, in most cases, the absolute values of Bias@C when using $P_{biased}$ are higher. These observations indicate that the LMs are affected by concept shortcuts within the prompt of ICL.

For utility performance, when changing from $P_{balanced}$ to $P_{biased}$, the average Acc@C and Acc@NoC decrease from 76.80\% to 76.42\% and from 76.37\% to 75.58\%, respectively, which means that spurious correlations harm utility performance regardless of the presence of concepts. 
For both Bias@C and accuracy, the relative change in the ICL scenario is less than the fine-tuning setting. We conjecture that a few exemplars in prompts make it hard to form a strong shortcut inside the LMs between conceptual contents and a specific label. 

\section{Mitigate Spurious Correlations}
\subsection{Mitigation via Rebalancing}
We consider two lines of existing data-centric work to mitigate spurious correlations: remove spurious components and rebalance the training dataset \cite{mccoy2019right, wang-etal-2022-identifying}. Since it is challenging to identify the conceptual contents in each sentence, we apply dataset rebalancing methods to mitigate the bias at the concept level.

We first downsample the dataset to achieve a balanced label distribution with respect to concept $c$, denoted as $\mathcal{D}^c_{down-bal}$. The shortcoming of the method is that, for a highly biased dataset, it filters out a large proportion of examples with the majority labels, leading to a sacrifice of the utility performance. To address this, we propose an upsampling method using ChatGPT to generate counterfactual examples with minority labels. Some concurrent work also demonstrates the effectiveness of synthetic data in mitigating bias \cite{evuru2024coda}. The resulting dataset is denoted as $\mathcal{D}^c_{up-bal}$.

Suppose that we need $\{a_0, \cdots, a_n\}$ number of examples for labels $\{0, \cdots, n\}$ to make a balanced subset for texts with concept $c$. We first sample $a_0$ to $a_n$ numbers of examples from texts with labels $0$ to $n$ but without concept $c$. Then we ask ChatGPT, $M_a$, to inject the concept $c$ into the selected texts while maintaining the sentiment or the answer to questions. Given the input text $x'$ without concept $c$, the injection prompt $P_i$ with the instruction, and the exemplars $h_c$ with concept $c$, the concept injection process is $x_c = M_a(P_i \| h_c \| x')$, where $x_c$ is the generated counterfactual for concept $c$ and input $x'$.
We iteratively generate the counterfactual input $x_c$ and add it into the dataset $\mathcal{D}_{ori}$ to form a balanced dataset $\mathcal{D}^c_{up-bal}$ with upsampling. To demonstrate the effectiveness of concept injection, we conduct a case study on a review in the Yelp dataset. As suggested in Table \ref{tab:case}, we inject the ``food'' concept into a review without this concept and observe that ChatGPT effectively injects the food concept, keeps other content unchanged, and maintains the sentiment of the review.

\begin{table}[t]
\resizebox{\columnwidth}{!}{%
\begin{tabular}{ll}
\toprule
Original & \begin{tabular}[c]{@{}l@{}}I was fairly disappointed with this zoo. \\ Signage was unclear. Many of the exhibits were on loan\end{tabular}                                                                      \\ \midrule
Counterfactual & \begin{tabular}[c]{@{}l@{}}I was fairly disappointed with this zoo. Signage was unclear. \\ Many of the exhibits were on loan. \\ \textbf{The food options consisted of a small cafe with limited choices.}\end{tabular} \\ \bottomrule
\end{tabular}
}
\caption{An example of the generated counterfactual data for concept ``food'' in the Yelp dataset. Text in bold is the generated input with the injected ``food'' concept.}
\label{tab:case}
\end{table}

We also propose a baseline method that masks out words highly associated with the concept. This method is used to verify whether balancing distributions of a few tokens removes conceptual shortcuts. We replace words with the top 10 PMI for each concept (word examples are in Table \ref{tab:top_pmi}) to the \textbf{[MASK]} token and name the masked dataset as $\mathcal{D}^c_{mask}$.

\begin{table*}[!htb]
\centering
\resizebox{\textwidth}{!}{%
\begin{tabular}{l|ccc|ccc|ccc|ccc}
\hline
\multirow{2}{*}{Data} & \multicolumn{3}{c|}{Amazon Shoe: Size}           & \multicolumn{3}{c|}{Amazon Shoe: Color}         & \multicolumn{3}{c|}{Amazon Shoe: Style}          & \multicolumn{3}{c}{IMDB: Acting}                \\
                        & Bias@C         & Acc@NoC        & Acc@C          & Bias@C        & Acc@NoC        & Acc@C          & Bias@C         & Acc@NoC        & Acc@C          & Bias@C        & Acc@NoC        & Acc@C          \\ \hline
$\mathcal{D}_{ori}$                  & 2.11           & \textbf{57.94} & 55.94 & \textbf{1.38} & 57.19          & 55.50          & 11.56          & 57.18          & 56.12          & 3.70          & 88.49          & 91.24          \\
$\mathcal{D}^c_{down-bal}$                 & 2.25           & 57.19          & 54.29          & 2.54          & 56.71          & 55.70          & 9.79           & \textbf{57.45} & 56.02          & 2.04          & 91.03          & 91.92          \\
$\mathcal{D}^c_{up-bal}$                   & \textbf{1.20}  & 56.72          & 55.64          & 2.06          & \textbf{57.44} & \textbf{57.01} & \textbf{9.16}  & 56.41          & \textbf{58.84} & \textbf{1.87} & \textbf{91.63} & \textbf{92.10} \\
$\mathcal{D}^c_{mask}$                 & 1.88           & 57.31          & \textbf{56.75} & 5.94          & 57.40          & 56.21          & 10.29          & 56.80          & 55.12          & 4.75          & 90.67          & 91.95          \\ \hline
                        & \multicolumn{3}{c|}{IMDB: Comedy}                & \multicolumn{3}{c|}{IMDB: Music}                & \multicolumn{3}{c|}{Yelp: Food}                  & \multicolumn{3}{c}{Yelp: Service}               \\
$\mathcal{D}_{ori}$                  & 2.51           & 91.62          & 91.85          & 12.07         & 91.55 & 88.93          & 3.42           & 93.86          & \textbf{97.09} & 2.44          & 93.33          & 97.23 \\
$\mathcal{D}^c_{down-bal}$                 & -0.37          & 90.86          & \textbf{92.88} & 7.71          & 91.03          & 92.29          & \textbf{-0.88} & 93.63          & 95.35          & 1.57          & 93.78          & 97.02          \\
$\mathcal{D}^c_{up-bal}$                   & \textbf{-0.32} & \textbf{91.65} & 92.74          & \textbf{4.05} & 90.19          & \textbf{92.80} & 2.86           & \textbf{94.15} & 96.85          & \textbf{0.39} & 93.84 & 97.04          \\
$\mathcal{D}^c_{mask}$                 & 1.08           & 90.41          & 92.56          & 8.28          & \textbf{91.88} & 90.52          & 2.02           & 93.52          & 96.45          & 1.02          & \textbf{94.67} & \textbf{97.41} \\
\hline
\end{tabular}
}
\caption{\textbf{Performance of multiple shortcut mitigation methods (downsampling, upsampling and token removal).} Upsampling method with the counterfactual generated data can obtain the best average effects in the aspects of reducing bias and increasing the utility performance. $\mathcal{D}_{ori}$ represents fine-tuning on the $\mathcal{D}_{ori}$ dataset.}
\label{tab:mitigation}
\end{table*}

\subsection{Results of Mitigation Methods}
\label{sec:fine-tuning_mitigate_results}
To evaluate the effectiveness of proposed methods, we select concepts with Bias@C greater than 1 in Table \ref{tab:finetune_main_results} and fine-tune on three de-biased datasets $\mathcal{D}^c_{down-bal}$, $\mathcal{D}^c_{up-bal}$, and $\mathcal{D}^c_{mask}$. We report results for DistilBERT in Table \ref{tab:mitigation} and \ref{tab:finetune_boolq} in Appendix B.

\noindent \textbf{Upsampling method reduces the bias and increases utility performance} \quad From Table \ref{tab:mitigation}, we observe that data rebalancing methods are effective in mitigating spurious correlations. For downsampling ($\mathcal{D}^c_{down-bal}$), it mitigates the mean absolute values of Bias@C from 4.90\% to 3.43\%, compared to trained on $\mathcal{D}_{ori}$. However, for utility performance, the downsampling obtains less accuracy in 4/8 cases for Acc@C and 5/8 cases for Acc@NoC, indicating that loss of data harms utility. For the upsampling method ($\mathcal{D}^c_{up-bal}$), the mean absolute values of Bias@C are effectively reduced from 4.90\% to 2.74\%. Furthermore, the average accuracy of Acc@C increases from 79. 24\% to 80. 38\%, and Acc@NoC is comparable. This observation suggests that adding counterfactual texts to rebalance the data can reduce spurious correlations in the concept level, and more data involved in the fine-tuning can boost the models' utility performance. 

Masking out associated tokens ($\mathcal{D}^c_{mask}$) can reduce spurious correlations in most cases, but cannot fully eliminate bias. 
This observation suggests that due to the various concept expressions, the learned concept shortcut in the model is not equivalent to the shortcut on a few tokens.
The utility performance of Acc@C is also lower than that of the proposed upsampling method in 6/8 comparisons.

In summary, among the three mitigation methods, adding the LLM-generated counterfactual inputs achieves the best performance in both the bias mitigation and utility aspects.
The same analysis on LLAMA2 models (Table \ref{tab:finetune_boolq} and \ref{tab:mitigation_llama} in Appendix \ref{sec:supplementary}) reveals similar patterns, which shows the generalizability of our methods.

\section{Related Work}
\paragraph{Robustness and Bias}
Current work on studying spurious correlations for LMs can be split into two categories: utilize the shortcuts during training or ICL. For shortcut learning in training, a series of works explores how models take shortcuts in the data for the causal or non-causal perspective \cite{tu2020empirical, sagawa2020investigation, geirhos2020shortcut, ribeiro2020beyond, kaushik2019learning, liu2024large, friedman2022finding} and which aspects of shortcuts will be taken for the predictions in different NLP tasks \cite{mccoy2019right, jia2017adversarial, lai-etal-2021-machine, zhao2018gender, niu2020evaluating, poliak2018hypothesis}, leading to low generalization in the out-of-distribution data or in the designed adversarial data.    

Due to the increasing development of LLM on ICL, researchers find that the design of the prompt significantly decides the LLM predictions \cite{brown2020language, gao2020making, liu2023pre, zhou2023scalable, schick2020exploiting}. Another line of work finds that LLMs are sensitive to a certain aspect of prompts and not robust when injecting adversarial triggers into prompt \cite{lu2021fantastically, zhao2021calibrate, tang2023large, si2023measuring, zheng2023large}. \citet{tang2023large} shows that LLMs use multiple types of shortcuts in the prompts, from letters to words to text style, and \citet{si2023measuring} find that LLMs exhibit clear feature biases under the unspecified prompts. 
Previous work also develops multiple methods to identify the topic or concept of text input \cite{li2024improving, abraham2022cebab, blei2003latent}.
However, our paper is the first to focus on assessing whether the models use shortcuts at the general concept level. 

\paragraph{Spurious Correlation Mitigation}
An increasing number of methods have attempted to mitigate spurious correlations in models caused by bias in the dataset \cite{chew2023understanding, clark2020learning, le2020adversarial, zhou-bansal-2020-towards, liu2021just, liu2023cat, zhu2023spottarget}, by data augmentation \cite{jin2020bert, alzantot2018generating, wang-etal-2022-identifying, minderer2020automatic}, data rebalancing \cite{mccoy2019right, sharma2018tackling, zellers2019hellaswag}, multi-task learning \cite{tu2020empirical}, and model ensembling or adding regularization \cite{utama2020mind,he2019unlearn, zhao2022investigating, liu2023c}.
To mitigate spurious correlations in a concept, we propose another data rebalancing method, which uses LLM to generate counterfactual sentences by injecting the concept and saves the human resource to compose them.

\section{Conclusions}
In this paper, we explore the spurious correlation at the general concept level in both fine-tuning and ICL settings. 
We find that LMs utilize the concept shortcut in training data (or in demonstrations) when inference on unseen data, and more biased training data (or prompts) lead to more biased predictions. To mitigate the learned shortcut, we propose a rebalancing method by adding counterfactual examples generated from ChatGPT to the training data, shown to be effective through extensive experiments. Our work indicates that 
LMs form strong spurious correlations on general concepts, encouraging future work to pay attention to unintended shortcut learning.

\section{Limitations}
Due to the limitation of the budget and the computation resource, we only fine-tuned the LLaMa2 7B model with the Lora method and used GPT3.5 for concept annotation. It could be interesting to fully fine-tune the LMs with a larger size. Moreover, in Section \ref{sec:experiments}, we find that the ChatGPT annotation of the concept label still achieves slightly lower accuracy than human annotators. We can use a more advanced model, such as GPT4 \cite{openai2023gpt4}, for annotation to increase the performance.

Our work focuses on five classification tasks. Three of them are binary classification tasks, and two are multiclass classification tasks. 
We apply the difference in accuracy for different groups (positive and negative) to measure bias at the concept level.  
Moreover, future work could extend our framework and generalize the measurement of concept bias to more complex tasks, such as the evaluation of LLM on QA tasks \cite{li2024panda} or even on tasks with the vision language model \cite{wang2024mementos, liu2023hallusionbench, wang2024enhancing, yue2023mmmu}.

For in-context learning, we observe that the concept bias in the demonstrations leads to larger spurious correlations. However, we also detect that the balanced prompts cannot fully eliminate the bias, and we do not provide a method to mitigate this inner spurious correlation in LMs. We leave that direction to future work.

\section*{Acknowledgments}

Zhou and Huang are supported by DARPA Transfer from Imprecise and Abstract Models to Autonomous Technologies (TIAMAT) 80321, National Science Foundation NSF-IIS-2147276 FAI, DOD-ONR-Office of Naval Research under award number N00014-22-1-2335, DOD-AFOSR-Air Force Office of Scientific Research under award number FA9550-23-1-0048, DOD-DARPA-Defense Advanced Research Projects Agency Guaranteeing AI Robustness against Deception (GARD) HR00112020007, Adobe, Capital One and JP Morgan faculty fellowships.
\bibliographystyle{acl_natbib}
\bibliography{anthology, custom}

\appendix
\section{Implementation Details}
\label{sec:implementation}

\subsection{Fine-tuning Experiments}

We use the DistilBERT and LLAMA2 model ~\cite{sanh2019distilbert, touvron2023llama} as our LMs for all of our fine-tuning experiments. For the DistilBERT model, we use AdamW as our optimizer with a learning rate of $2\mathrm{e}{-5}$ and a weight decay of $0.01$ with linear scheduler, batch size of $16$, and trained for $3$ epochs. For the LLAMA2 model, we use AdamW as our optimizer with a learning rate of $2\mathrm{e}{-4}$, batch size of $32$, warm-up ratio of 0.03, and trained for $3$ epochs. We base our implementation on the Pytorch\footnote{\url{https://pytorch.org/}}, Huggingface transformer\footnote{\url{https://huggingface.co/}} frameworks, and the LLAMA2 weights from Meta\footnote{\url{https://ai.meta.com/llama/}}.

\subsection{ICL Setup}
We utilize greedy search in decoding for all ICL experiments and counterfactual data generation, except for the annotation of concepts for each text, where we use stochastic temperature sampling with the temperature value 0.7 to obtain diverse answers. The template of the prompts for the ICL experiments, concept annotations and counterfactual data generations are suggested in Table \ref{table:icl_concept_prompt}, Table \ref{table:icl_exp_prompt} and Table \ref{table:counterfactual_prompt}.

We call the gpt-3.5-turbo (4k) function from OpenAI to generate the concept labels, ICL experiments and concept injection. The price of this API is \$0.0015 / 1K tokens for inputs and \$0.002 / 1K tokens for output. The total expenditure on API usage is about \$ 300.00, including preliminary exploration.

\section{Prompt Details and Supplementary Results}
\label{sec:supplementary}
In Table \ref{tab:finetune_boolq}, we perform the same analysis on the BoolQ question and answering dataset for all experiments (ICL and fine-tuning) in Section \ref{sec:main_result}.
In Table \ref{tab:finetune_main_results_llama} and Table \ref{tab:mitigation_llama}, we repeat the experiments for fine-tuning LLAMA2 7B models for Section \ref{sec:fine-tuning_results} and \ref{sec:fine-tuning_mitigate_results}. In Table \ref{table:icl_concept_prompt}, \ref{table:icl_exp_prompt}, and \ref{table:counterfactual_prompt}, we present the details of the prompts in the annotation of the concept, the ICL experiments, and the countergactual sentence generation.

\begin{table}[!htbp]
\centering
\resizebox{\linewidth}{!}{%
\begin{tabular}{l|ccc|ccc}
\hline
\multirow{2}{*}{Distilbert} & \multicolumn{3}{c|}{BoolQ Country (pos \textless neg)}        & \multicolumn{3}{c}{BoolQ History (pos \textgreater neg)} \\
                        & Bias@C            & Acc@NoC          & Acc@C            & Bias@C            & Acc@NoC            & Acc@C \\ \hline
Trained on $\mathcal{D}_{ori}$      &  \textbf{-1.22}   &  \textbf{61.23} &  \textbf{60.95}          & 5.21    &  59.93    & 57.85     \\
Trained on $\mathcal{D}^c_{biased}$       &   -18.90  &    55.92        &  55.46  & 50.63     &  57.37    & 55.22 \\
Trained on $\mathcal{D}^c_{down-bal}$       &  4.84   &    57.93        &  58.79  & -8.95     &  59.87    & 58.60 \\
Trained on $\mathcal{D}^c_{up-bal}$       &  2.45   &   59.54         & 59.74   &  \textbf{-0.85}   &  60.13    & 59.70 \\
Trained on $\mathcal{D}^c_{mask}$       &  2.90   &     60.71       & 59.69   &  -1.55    &   \textbf{60.94}   & \textbf{60.84} \\ \hline
LLAMA2 & Bias@C            & Acc@NoC          & Acc@C            & Bias@C            & Acc@NoC            & Acc@C \\ \hline
Trained on $\mathcal{D}_{ori}$       &  -9.72   &    67.78        &  72.68  &   \textbf{0.82}   &  78.87    & 80.43 \\
Trained on $\mathcal{D}^c_{biased}$       &  -9.67   &     67.23       &  70.86  &  2.73    &   75.86   & 78.79 \\
Trained on $\mathcal{D}^c_{down-bal}$       &  -10.18   &  77.11          &  \textbf{81.61}  &  1.20    &   76.11   &  77.64 \\
Trained on $\mathcal{D}^c_{up-bal}$   &   \textbf{-7.81}  &   77.30         &  78.16      &  -3.23    &   \textbf{80.62}   & \textbf{82.40} \\
Trained on $\mathcal{D}^c_{mask}$       &  -10.55   &   \textbf{77.53}         &  79.10  &  -7.73    &  78.91    & 
 80.73 \\ \hline
GPT3.5 ICL & Bias@C            & Acc@NoC          & Acc@C            & Bias@C            & Acc@NoC            & Acc@C \\ \hline
ICL with $P_{balanced}$      & \textbf{-1.80}    &   81.98         &  \textbf{83.99}  &  \textbf{-3.88}    &  \textbf{83.12}    & \textbf{83.90} \\
ICL with $P_{biased}$       &  -2.90   &    \textbf{82.93}        & 83.11   & -3.93     &  82.03    & 82.82 \\ \hline
\end{tabular}
}
\caption{Fine-tuning and ICL performance for all experiments in Section \ref{sec:main_result} on BoolQ dataset of DistilBert, LLAMA2 (fine-tuning) and GPT3.5 (ICL) models. The smaller absolute values of Bias@C (smaller bias) and larger values of Acc are in bold.}
\label{tab:finetune_boolq}
\end{table}

\begin{table*}[!htbp]
\centering
\resizebox{\textwidth}{!}{%
\begin{tabular}{l|ccc|ccc|ccc}
\hline
\multirow{2}{*}{Method} & \multicolumn{3}{c|}{AS Size (pos \textless neg)}        & \multicolumn{3}{c|}{AS Color (pos \textgreater neg)}      & \multicolumn{3}{c}{AS Style (pos \textgreater neg)}       \\
                        & Bias@C            & Acc@NoC          & Acc@C            & Bias@C            & Acc@NoC            & Acc@C            & Bias@C             & Acc@NoC           & Acc@C            \\ \hline
Trained on $\mathcal{D}_{ori}$       & \textbf{-1.46}    & 59.46            & \textbf{57.24}   & \textbf{3.87}     & \textbf{59.62}     & 59.07            & \textbf{16.01}     & 59.54             & \textbf{58.89}   \\
Trained on $\mathcal{D}^c_{biased}$    & -7.69             & \textbf{59.57}   & 51.86            & 7.93              & 58.54              & \textbf{59.43}   & 17.31              & \textbf{59.91}    & 58.14            \\ \hline
\multirow{2}{*}{Method} & \multicolumn{3}{c|}{IMDB Acting (pos \textgreater neg)} & \multicolumn{3}{c|}{IMDB Comedy (pos \textgreater neg)}   & \multicolumn{3}{c}{IMDB Music (pos \textgreater neg)}     \\
                        & Bias@C            & Acc@NoC          & Acc@C            & Bias@C            & Acc@NoC            & Acc@C            & Bias@C             & Acc@NoC           & Acc@C            \\ \hline
Trained on $\mathcal{D}_{ori}$       & \textbf{3.51}     & \textbf{95.87}   & \textbf{97.55}   & 1.23              & 96.30              & \textbf{97.55}   & \textbf{5.35}      & 96.84             & 95.69            \\
Trained on $\mathcal{D}^c_{biased}$    & 4.61              & 95.73            & 97.54            & \textbf{0.40}     & \textbf{96.79}     & 97.19            & 6.65               & \textbf{96.88}    & \textbf{96.24}   \\ \hline
\multirow{2}{*}{Method} & \multicolumn{3}{c|}{Yelp Food (pos \textgreater neg)}   & \multicolumn{3}{c|}{Yelp Service (pos \textless neg)}     & \multicolumn{3}{c}{Yelp Price (pos \textgreater neg)}     \\
                        & Bias@C            & Acc@NoC          & Acc@C            & Bias@C            & Acc@NoC            & Acc@C            & Bias@C             & Acc@NoC           & Acc@C            \\ \hline
Trained on $\mathcal{D}_{ori}$       & \textbf{2.62}     & \textbf{98.30}   & \textbf{98.80}   & 1.41              & \textbf{97.79}     & \textbf{99.24}   & \textbf{0.32}      & \textbf{98.59}    & \textbf{98.52}   \\
Trained on $\mathcal{D}^c_{biased}$    & 2.64              & 98.23            & 98.80            & \textbf{1.11}     & 97.78              & 99.19            & 0.78               & 98.37             & 98.37            \\ \hline
\multirow{2}{*}{Method} & \multicolumn{3}{c|}{CeBaB Food (pos \textgreater neg)}  & \multicolumn{3}{c|}{CeBaB Service (pos \textgreater neg)} & \multicolumn{3}{c}{CeBaB Ambiance (pos \textgreater neg)} \\
                        & Bias@C            & Acc@NoC          & Acc@C            & Bias@C            & Acc@NoC            & Acc@C            & Bias@C             & Acc@NoC           & Acc@C            \\ \hline
Trained on $\mathcal{D}_{ori}$       & \textbf{3.04}     & \textbf{74.01}   & \textbf{75.57}   & -3.44             & \textbf{69.58}     & \textbf{75.21}   & \textbf{0.41}      & \textbf{73.05}    & \textbf{75.60}   \\
Trained on $\mathcal{D}^c_{biased}$    & 3.67              & 61.96            & 65.09            & \textbf{3.12}     & 65.05              & 67.22            & 1.35               & 68.97             & 73.63            \\ \hline
\end{tabular}
}
\caption{\textbf{Model fine-tuning performance with training on original dataset and concept biased dataset for LLAMA2 fine-tuning.} pos > neg: The number of positive texts is larger than the number of negative texts in the original data and in biased dataset, all texts containing this concept are positive, and vice versa for ``pos < neg''. The smaller absolute values of Bias@C (smaller bias) and larger values of Acc are in bold.}
\label{tab:finetune_main_results_llama}
\end{table*}

\begin{table*}[!htbp]
\centering
\resizebox{\textwidth}{!}{%
\begin{tabular}{l|ccc|ccc|ccc|ccc|ccc}
\hline
\multirow{2}{*}{Method} & \multicolumn{3}{c|}{Amazon Shoe: Size}           & \multicolumn{3}{c|}{Amazon Shoe: Color}          & \multicolumn{3}{c|}{Amazon Shoe: Style}          & \multicolumn{3}{c|}{IMDB: Acting}                & \multicolumn{3}{c}{CeBaB: Food}                                                      \\
                        & Bias@C         & Acc@NoC        & Acc@C          & Bias@C         & Acc@NoC        & Acc@C          & Bias@C         & Acc@NoC        & Acc@C          & Bias@C         & Acc@NoC        & Acc@C          & \multicolumn{1}{l}{Bias@C} & \multicolumn{1}{l}{Acc@NoC} & \multicolumn{1}{l}{Acc@C} \\ \hline
$\mathcal{D}_{ori}$                   & -1.46          & 59.46          & 57.24          & 3.87           & 59.62          & 59.07          & 16.01          & \textbf{59.54} & 58.89          & 3.51           & 95.87          & 97.55          & 3.04                       & \textbf{74.01}              & \textbf{75.57}            \\
$\mathcal{D}^c_{down-bal}$                 & 2.88           & 59.17          & \textbf{57.93} & 4.10           & 59.57          & 58.48          & 11.62          & 58.13          & \textbf{61.83} & 2.28           & \textbf{96.14} & 97.59          & 2.16                       & 68.86                       & 71.09                     \\
$\mathcal{D}^c_{up-bal}$                   & \textbf{-0.62} & \textbf{59.94} & 56.53          & \textbf{-2.01} & \textbf{59.92} & \textbf{60.90} & \textbf{11.37} & 59.32          & 60.43          & \textbf{1.91}  & 96.12          & 97.72          & 1.83                       & 70.71                       & 74.75                     \\
$\mathcal{D}^c_{mask}$                 & -2.41          & 59.65          & 53.65          & 4.12           & 58.23          & 58.33          & 13.33          & 59.12          & 59.48          & 2.50           & 96.02          & \textbf{97.73} & \textbf{0.02}              & 73.45                       & 73.42                     \\ \hline
                        & \multicolumn{3}{c|}{IMDB: Comedy}                & \multicolumn{3}{c|}{IMDB: Music}                 & \multicolumn{3}{c|}{Yelp: Food}                  & \multicolumn{3}{c|}{Yelp: Service}               & \multicolumn{3}{c}{CeBaB: Service}                                                   \\
$\mathcal{D}_{ori}$                   & 1.23           & 96.30          & 97.55          & 5.35           & 96.84          & 95.69          & 2.62           & 98.30          & 98.80          & 1.41           & 97.79          & 99.24          & -3.44                      & 69.58                       & 75.21                     \\
$\mathcal{D}^c_{down-bal}$                 & 0.89           & 96.32          & 97.32          & 8.56           & \textbf{97.27} & 95.12          & 4.39           & 98.18          & 98.92          & -0.35 & \textbf{98.00}          & 99.20          & -1.86                      & \textbf{70.44}              & 74.84                     \\
$\mathcal{D}^c_{up-bal}$                   & 0.77           & \textbf{97.59} & \textbf{97.74} & 8.01           & 96.78          & 94.57          & \textbf{1.83}  & 97.91          & \textbf{99.04} & 0.41           & 97.62          & \textbf{99.30} & \textbf{0.21}              & 70.25                       & \textbf{75.88}            \\
$\mathcal{D}^c_{mask}$                 & \textbf{0.53}  & 96.89          & 97.18          & \textbf{2.68}  & 96.98          & \textbf{96.71} & 3.18           & \textbf{98.36} & 98.61          & \textbf{0.08}               & 97.96      & 98.82      & -0.90                      & 69.83                       & 75.05                     \\ \hline
\end{tabular}
}
\caption{\textbf{Performance of multiple shortcut mitigation methods (downsampling, upsampling and token removal) for LLAMA2 fine-tuning.} Upsampling method with the counterfactual generated data can obtain the best average effects in the aspects of reducing bias and increasing the utility performance.}
\label{tab:mitigation_llama}
\end{table*}

\begin{table*}[!htbp]
    \centering
    \begin{tabular}{m{40em}}
        \hline
        I will provide you 5 reviews in \textcolor{red}{\{dataset name\}} dataset. Please find the concepts explicitly mentioned in this review only from the set with three concepts:  \textcolor{red}{\{candidate concepts\}}. Do not include other concepts. If you can not find any of these concepts in the concept set, please annotate this review with ``none''. Wrap your answer for a review in a word sequence separated by the comma and for each answer, start with a new line with an index. \\
        Here are a few examples:  \\
        \textcolor{red}{\{demonstrations\}} \\
        The output is: \\
        \textcolor{red}{\{output concepts\}}
        \\
        Here is the review list of 5 OpenTable reviews: \\
        \textcolor{red}{\{text lists\}} \\
        The output is: \\
        \hline
        
    \end{tabular}
    \caption{\label{table:icl_concept_prompt} Prompt $P_a$ for concept annotation in all datasets. \textcolor{red}{\{dataset name\}} and \textcolor{red}{\{candidate concepts\}} are placeholders to put the name of dataset and the candidate concepts. For example, for Amazon shoe dataset, they are ``Amazon shoe'' and ``size, color, and style''. \textcolor{red}{\{demonstrations\}} and \textcolor{red}{\{output concepts\}} are placeholders to add five demonstrations with provided ground-truth concept labels. \textcolor{red}{\{text lists\}} is a placeholder to add the text to be annotated.}
\end{table*}

\begin{table*}[ht]
    \begin{subtable}{1\textwidth}
    \centering
    \begin{tabular}{m{40em}}
        \hline
         Given a review, you need to predict whether the sentiment of the review is positive or negative.
        Here are the examples: \\
        Review: \textcolor{red}{\{review 1\}} Sentiment label: \textcolor{red}{\{label 1\}}\\
        Review: \textcolor{red}{\{review 2\}} Sentiment label: \textcolor{red}{\{label 2\}}\\
        Review: \textcolor{red}{\{review 3\}} Sentiment label: \textcolor{red}{\{label 3\}}\\
        Review: \textcolor{red}{\{review 4\}} Sentiment label: \textcolor{red}{\{label 4\}}\\
        Review: \textcolor{red}{\{review 5\}} Sentiment label: \textcolor{red}{\{label 5\}}\\
        Review: \textcolor{red}{\{review 6\}} Sentiment label: \textcolor{red}{\{label 6\}}\\
        Review: \textcolor{red}{\{review 7\}} Sentiment label: \textcolor{red}{\{label 7\}}\\
        Review: \textcolor{red}{\{review 8\}} Sentiment label: \textcolor{red}{\{label 8\}}\\
        Here is the review to predict sentiment: \\
        Review: \textcolor{red}{\{$x_{test}$\}} Sentiment label: \\
        \hline
        
    \end{tabular}
    \caption{Prompt $P_{balanced}$ or $P_{biased}$ for IMDB and Yelp dataset.}
    \end{subtable}
    \bigskip
    \begin{subtable}{1\textwidth}
    \centering
    \begin{tabular}{m{40em}}
        \hline
        Given a review, you need to predict whether the sentiment label of the review from 0 to 4, total 5 classes. Label 0 represents the most negative review and Label 4 represents the most positive review.
        Here are the examples: \\
        Review: \textcolor{red}{\{review 1\}} Sentiment label: \textcolor{red}{\{label 1\}}\\
        Review: \textcolor{red}{\{review 2\}} Sentiment label: \textcolor{red}{\{label 2\}}\\
        Review: \textcolor{red}{\{review 3\}} Sentiment label: \textcolor{red}{\{label 3\}}\\
        Review: \textcolor{red}{\{review 4\}} Sentiment label: \textcolor{red}{\{label 4\}}\\
        Review: \textcolor{red}{\{review 5\}} Sentiment label: \textcolor{red}{\{label 5\}}\\
        Review: \textcolor{red}{\{review 6\}} Sentiment label: \textcolor{red}{\{label 6\}}\\
        Review: \textcolor{red}{\{review 7\}} Sentiment label: \textcolor{red}{\{label 7\}}\\
        Review: \textcolor{red}{\{review 8\}} Sentiment label: \textcolor{red}{\{label 8\}}\\
        Here is the review to predict sentiment: \\
        Review: \textcolor{red}{\{$x_{test}$\}} Sentiment label: \\
        \hline
        
    \end{tabular}
    \caption{Prompt $P_{balanced}$ or $P_{biased}$ for CeBaB and Amazon shoe dataset.}
    \end{subtable}
    \caption{\label{table:icl_exp_prompt} Prompt $P_{balanced}$ or $P_{biased}$ for the ICL experiments for all datasets. \textcolor{red}{\{review\}} and \textcolor{red}{\{label\}} is a placeholder to add 8 demonstrations with provided ground-truth sentiment labels for each dataset. \textcolor{red}{\{$x_{test}$\}} is the place to insert the predicted text.}
\end{table*}

\begin{table*}[!htb]
    \centering
    \begin{tabular}{m{40em}}
        \hline
        Here are 5 exemplars with the \textcolor{red}{\{concept\}} concept: \\
        \textcolor{red}{\{texts with concept\}} \\
        Here are another 5 exemplars without the \textcolor{red}{\{concept\}} concept: \\
        \textcolor{red}{\{texts without concept\}} \\
        Please inject the “{concept}” concept into a statement and maintain the sentiment level of this statement. \\
        The statement is: \\
        \textcolor{red}{\{text for counterfactual\}} \\
        The output statement with the \textcolor{red}{\{concept\}}  concept is: \\
        \hline
        
    \end{tabular}
    \caption{\label{table:counterfactual_prompt} Prompt $P_i$ for counterfactual data generation in all datasets. \textcolor{red}{\{concept\}} are a placeholder to put the concept for generating the counterfactual data. \textcolor{red}{\{texts with concept\}} and \textcolor{red}{\{texts without concept\}} are placeholders to add five demonstrations with or without the concepts. \textcolor{red}{\text for counterfactual\}} is a placeholder to add the text to make the counterfactual in the concept level.}
\end{table*}

\end{document}